# Spatiotemporal Facial Action Unit Detection using Twin Cycle Autoencoders for Driver Monitoring

Sai Sidharth D

*sidharthsai.d@gmail.com*

***Abstract—*** *Driver monitoring systems (DMS) increasingly rely on facial cues to infer drowsiness, distraction, and cognitive load in real time. Facial Action Units (AUs), grounded in the Facial Action Coding System (FACS), provide an objective and interpretable representation of such states, but their automatic detection in the driving context is complicated by low and variable illumination, partial occlusion, head-pose variation, and the subtlety and short duration of relevant AU activations. Existing AU detectors largely treat spatial appearance and temporal dynamics separately, limiting their ability to exploit self-supervisory signal from abundant unlabeled driving video. We propose the Twin Cycle Autoencoder (TCA), a spatiotemporal architecture composed of two coupled cycle-consistent autoencoder branches: a Spatial Cycle Autoencoder that disentangles AU-relevant appearance from identity through image-level cycle consistency, and a Temporal Cycle Autoencoder that enforces forward-backward consistency over latent AU trajectories to capture onset-apex-offset dynamics. The two branches are coupled through a cross-branch latent alignment loss and fused via an attention module before multi-label AU classification. We evaluate TCA on the DISFA and BP4D benchmarks and on an in-cabin naturalistic driving dataset, and observe consistent improvements over CNN-RNN, 3D-CNN, and graph-based AU baselines, particularly for low-intensity and rapidly transitioning AUs relevant to fatigue (AU45, AU43) and yawning (AU26). We further show the model sustains real-time throughput on an embedded Jetson Xavier NX platform, supporting its use in production-grade advanced driver assistance systems (ADAS).*

## I. Introduction

Road safety statistics consistently identify driver drowsiness and inattention as leading contributors to fatal and near-fatal crashes, motivating the integration of camera-based Driver Monitoring Systems (DMS) into modern vehicles. While early DMS relied on coarse indicators such as steering-wheel behavior, lane-keeping deviation, or eye-closure percentage (PERCLOS), recent systems increasingly leverage fine-grained facial cues that correlate with underlying physiological and cognitive states. Facial Action Units (AUs), defined by the Facial Action Coding System (FACS), decompose facial expressions into anatomically grounded, independently interpretable muscle movements, offering a principled intermediate representation between raw pixels and high-level affect or fatigue labels.

Detecting AUs reliably inside a vehicle cabin is substantially harder than in controlled laboratory settings. In-cabin cameras must contend with strong and rapidly changing illumination (sun glare, tunnel transitions, nighttime infrared imaging), partial occlusion from sunglasses, hands, or masks, extreme head-pose variation during shoulder checks, and AUs that are individually low in intensity and brief in duration, such as micro-expressions of fatigue. Moreover, most publicly available AU-annotated corpora (e.g., DISFA, BP4D) are recorded in frontal, well-lit, non-driving conditions, so models trained purely on such data generalize poorly to naturalistic driving footage, while densely AU-labeled driving data remains scarce and expensive to collect.

A further limitation of the AU detection literature is architectural: the majority of deep learning approaches either process frames independently with strong spatial backbones and attention modules, discarding temporal context, or bolt a generic recurrent or 3D-convolutional temporal head onto spatial features without an explicit mechanism to regularize the temporal representation using unlabeled video. Cycle-consistency, popularized in unpaired image translation and later applied to temporal alignment of unlabeled video, provides such a mechanism: it allows a network to be trained, in part, without AU labels by requiring that transformations be invertible in a well-defined sense. We hypothesize that a twin architecture, applying cycle consistency independently in the spatial (appearance) and temporal (dynamics) domains and coupling the two through a shared latent space, yields AU representations that are both identity-invariant and temporally coherent, and that this is particularly beneficial in the label-scarce, high-variability driving setting.

This paper makes the following contributions:

1) We propose the Twin Cycle Autoencoder (TCA), a spatiotemporal architecture comprising a Spatial Cycle Autoencoder (SCA) for AU-identity disentanglement and a Temporal Cycle Autoencoder (TCAE-T) for forward-backward temporal consistency over AU latent trajectories.

2) We introduce a cross-branch latent alignment loss that couples spatial and temporal representations, together with an attention-based fusion module for multi-label AU classification.

3) We design a self-supervised pretraining strategy that exploits large volumes of unlabeled in-cabin driving video via the cycle-consistency objectives alone, prior to AU-supervised fine-tuning.

4) We report extensive experiments on DISFA, BP4D, and a naturalistic driving dataset, including ablations and an embedded real-time inference analysis on a Jetson Xavier NX platform.

## II. Related Work

### A. Facial Action Unit Detection

Early AU detection systems used handcrafted appearance and geometric features (e.g., Gabor wavelets, HOG, and Active Appearance Models) with SVM or boosting classifiers [1]. Deep convolutional networks subsequently improved AU detection accuracy by learning discriminative features directly from data, with region-based and attention-based architectures such as JAA-Net jointly optimizing AU detection with facial landmark localization to focus on AU-relevant regions [2]. More recent work models statistical or semantic dependencies between AUs using graph neural networks, improving performance on co-occurring AUs [3]. However, these methods are predominantly frame-based and do not explicitly exploit temporal continuity.

### B. Spatiotemporal Deep Learning for Facial Analysis

Temporal modeling for facial video has been approached with recurrent networks (LSTM, ConvLSTM) stacked on CNN features [4], 3D convolutions that jointly learn spatial and temporal filters [5], and, more recently, video transformers with spatiotemporal attention. While effective, these architectures generally require large labeled datasets to learn stable temporal representations and offer no explicit self-supervised regularization signal for the temporal dimension when labels are scarce, a common situation for AU-annotated driving data.

### C. Autoencoders and Cycle Consistency

Cycle-consistent adversarial networks (CycleGAN) demonstrated that requiring a transformation to be approximately invertible enables learning from unpaired data without direct supervision [6]. This idea has since been extended to temporal alignment of video via Temporal Cycle Consistency (TCC) learning, which aligns frames across sequences of the same action without labels [7]. Variational and adversarial autoencoders have separately been used to disentangle identity from expression in the spatial domain [8]. TCA unifies these two lines of work, applying cycle consistency separately but jointly in the spatial and temporal domains for the AU detection task.

### D. Driver Monitoring Systems

Commercial and research DMS have historically relied on eye-closure and gaze metrics such as PERCLOS [9] and head-pose/gaze estimation [10] to infer drowsiness or distraction. Multimodal approaches combine facial video with physiological or vehicle-dynamics signals [11]. Direct AU-based drowsiness modeling remains comparatively underexplored, largely due to the absence of large, densely AU-labeled naturalistic driving corpora, motivating the self-supervised pretraining component of the proposed method.

## III. Proposed Method

### A. Problem Formulation

Given an input video clip $X = \{I_{t-k}, ..., I_t\}$ of k+1 consecutive cropped and aligned face frames, the goal is to predict, for the final frame $I_t$, a multi-label AU occurrence vector $\hat{y} \in \{0,1\}^N$ (and optionally an intensity vector for a subset of AUs), where N is the number of target AUs (N = 12 for the driving-relevant AU set used in our experiments). The model is trained using a combination of a small labeled set $D_L = \{(X_i, y_i)\}$ and a substantially larger unlabeled set $D_U = \{X_j\}$ of driving video without AU annotation.

### B. Overall Architecture

TCA (Fig. 1) consists of two coupled branches operating at different granularities. The Spatial Cycle Autoencoder (SCA) operates independently on each frame $I_t$, encoding it with a convolutional encoder $E_S$ (ResNet-18 backbone, pretrained on face-recognition data) into an AU-relevant latent code $z_t^S \in R^d$, discarding identity-specific information. A decoder $D_S$ reconstructs the frame from $z_t^S$ together with an auxiliary identity code, and a second, AU-manipulated latent code (obtained by perturbing $z_t^S$ along learned AU directions) is re-encoded and compared against the manipulation target, forming the spatial cycle: image → latent → image → latent.

The Temporal Cycle Autoencoder (TCAE-T) operates on the sequence of spatial latent codes $Z^S = \{z_{t-k}^S, ..., z_t^S\}$. A temporal encoder $E_T$ (a bidirectional GRU or lightweight Transformer encoder, compared in Section VII) maps this sequence to a temporal latent trajectory, and a temporal decoder $D_T$ reconstructs the sequence in reverse order from the trajectory; the discrepancy between the reconstructed and true reversed sequence forms the temporal cycle-consistency loss, which encourages the trajectory to encode dynamics that are informative in both time directions, analogous to forward-backward consistency used in TCC-style alignment but applied at the level of an explicit generative decoder rather than soft nearest-neighbor alignment.

### C. Cross-Branch Latent Coupling and Fusion

The two branches are coupled through a cross-branch latent alignment loss that minimizes the distance between the temporal trajectory's instantaneous state at time t and the spatial latent code $z_t^S$, projected into a shared embedding space via a small MLP. This coupling discourages the temporal branch from drifting away from AU-grounded spatial semantics while still allowing it to model dynamics not visible in a single frame. Formally, let $h_t$ denote the hidden state produced by the temporal encoder $E_T$ at time t, and let $g(\cdot)$ denote the learned MLP projection head mapping the temporal state into the shared d-dimensional embedding space used by the spatial branch. The cross-branch latent alignment loss over a clip of length T is defined in Eq. (1):

$$L_{align} = (1/T)\ \Sigma_{t=1}^{T}\ d(\ g(h_t), z_t^S\ ) \quad (1)$$

where $d(\cdot,\cdot)$ is a distance function measured in the shared embedding space. We consider two standard choices: the squared Euclidean distance, $d_{MSE}(a,b) = \|a - b\|_2^2$, and the cosine distance, $d_{cos}(a,b) = 1 - (a\cdot b)/(\|a\|\|b\|)$. In our experiments we adopt the cosine distance as the default for $L_{align}$, since embedding direction, rather than magnitude, is the more meaningful carrier of AU-relevant semantic content across the two branches, and cosine distance was empirically more stable during self-supervised pretraining, where the unconstrained magnitude of $h_t$ can otherwise dominate the squared-Euclidean term early in training before the temporal encoder's output scale stabilizes; a direct comparison between the two distance choices is reported as part of the ablation in Section VII.

The aligned spatial and temporal embeddings are combined using a lightweight multi-head attention fusion module, producing a single spatiotemporal descriptor $f_t$ that is passed to a fully connected AU classification head with a sigmoid activation per AU, trained with weighted binary cross-entropy to address class imbalance (many AUs are rare/negative in naturalistic driving).

### *D. Loss Functions*

The total training objective is a weighted sum of five terms, given in Eq. (2):

$$L = \lambda_1 L_{rec}^S + \lambda_2 L_{cyc}^S + \lambda_3 L_{cyc}^T + \lambda_4 L_{align} + \lambda_5 L_{cls} \quad (2)$$

where $L_{rec}^S$ is the spatial pixel/perceptual reconstruction loss, $L_{cyc}^S$ and $L_{cyc}^T$ are the spatial and temporal cycle-consistency losses described above, $L_{align}$ is the cross-branch latent alignment loss defined in Eq. (1), and $L_{cls}$ is the weighted multi-label AU classification loss. During self-supervised pretraining on $D_U$, $\lambda 5 = 0$ since AU labels are unavailable; $\lambda 5$ is annealed to its full value during supervised fine-tuning on $D_L$. Weights $\lambda 1$–$\lambda 5$ are tuned on a held-out validation split (Section V).

### *E. Implementation Details*

The spatial encoder is a ResNet-18 truncated after the penultimate residual block, producing $z^S \in R^{256}$. The temporal encoder uses either a 2-layer bidirectional GRU with 256 hidden units per direction (512-dimensional concatenated output) or a 4-layer Transformer encoder with 8 attention heads and a 256-dimensional model width (compared in Section VII); both operate on 16-frame clips sampled at 10 fps ($\approx$ 1.6 s temporal context), which we found sufficient to capture full AU onset-apex-offset cycles for the fast AUs relevant to drowsiness (blink, yawn onset) while remaining tractable for embedded inference.

The alignment projection head $g(\cdot)$ is implemented as a lightweight 2-layer MLP that maps the temporal encoder's per-timestep hidden state $h_t$ into the shared 256-dimensional embedding space occupied by the spatial latent code $z_t^S$. For the GRU temporal encoder, whose bidirectional output is 512-dimensional, $g(\cdot)$ has the configuration Linear(512→384) → ReLU → Dropout(0.1) → Linear(384→256) → L2-normalize; for the Transformer temporal encoder, whose output is already 256-dimensional, a narrower head is used, Linear(256→256) → ReLU → Dropout(0.1) → Linear(256→256) → L2-normalize, to avoid an unnecessary bottleneck-then-expand pattern. The spatial latent $z_t^S$ is likewise passed through an L2-normalization (no additional trainable layers, since it is already the target embedding space) before the cosine distance in Eq. (1) is computed, ensuring both branches are compared on the unit hypersphere rather than in raw, differently-scaled coordinate systems. This shared 256-dimensional target space was chosen to match the spatial encoder's native latent width, avoiding an additional projection on the spatial branch that could itself discard AU-relevant information.

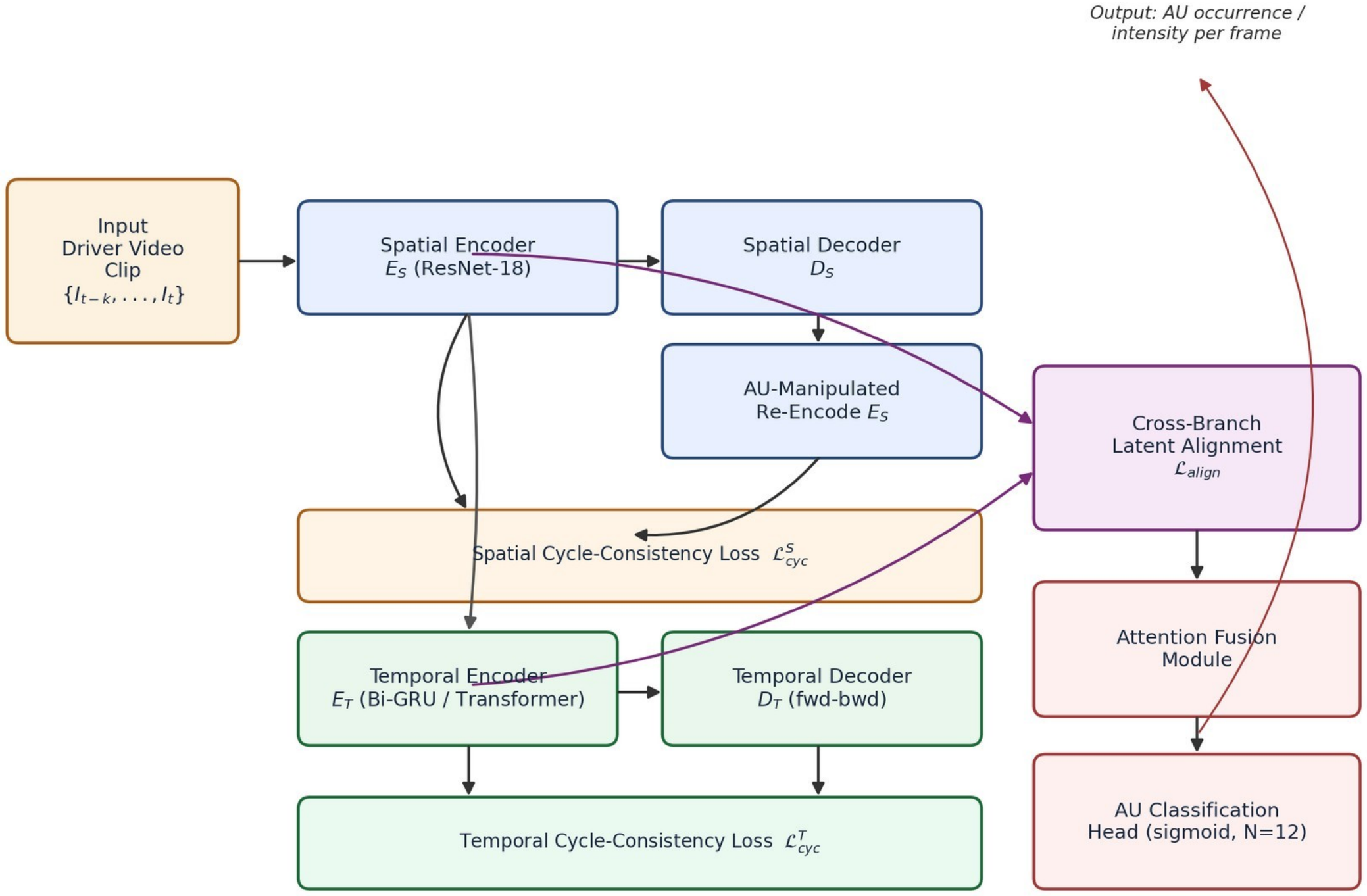

*Fig. 1. Overview of the Twin Cycle Autoencoder (TCA): coupled Spatial and Temporal Cycle Autoencoder branches, cross-branch latent alignment, attention fusion, and the AU classification head.*

## IV. Dataset and Preprocessing

### A. Benchmark Datasets

DISFA contains spontaneous facial expressions from 27 subjects recorded under controlled frontal conditions, annotated with frame-level intensity (0–5) for 12 AUs [12]. BP4D comprises 41 subjects performing 8 emotion-elicitation tasks with occurrence labels for 12 core AUs [13]. Following common practice, AU occurrence is derived from DISFA intensities using a threshold of ≥ 2, and we report results on the standard 3-fold subject-independent protocol for both datasets.

### B. Naturalistic Driving Dataset

To evaluate in-cabin generalization, we use a naturalistic driving-simulator dataset collected under near-infrared (NIR) illumination from 45 participants across daytime and simulated nighttime conditions, including sessions with sunglasses and partial hand occlusion. A subset of 9 driving-relevant AUs (AU4, AU6, AU7, AU9, AU12, AU24, AU26, AU43, AU45) was annotated by two certified FACS coders on 15% of frames, with the remainder left unlabeled for self-supervised pretraining, yielding approximately 38 hours of unlabeled video and 5.7 hours of AU-labeled video.

### C. Preprocessing

For all datasets, faces are detected and aligned using RetinaFace, with 68-point landmarks refined via a lightweight landmark regressor for the NIR driving footage. Faces are cropped to 224×224, and, for the driving dataset, single-channel NIR frames are replicated across three channels prior to normalization. Standard augmentations (random crop, horizontal flip disabled to preserve AU asymmetry semantics, brightness/contrast jitter, and synthetic occlusion patches) are applied during supervised fine-tuning only.

## V. Experimental Setup

### A. Baselines and Metrics

We compare TCA against: (i) JAA-Net [2], a strong frame-based spatial baseline; (ii) ConvLSTM-AU, a ConvLSTM temporal extension of a ResNet backbone; (iii) a 3D-CNN (I3D-style) AU detector [5]; and (iv) ME-GraphAU, a graph-based AU relational model [3]. Performance is reported as the mean F1 score (F1) across AUs, following standard AU detection evaluation practice, alongside AUC-ROC.

### B. Training Configuration

Models are implemented in PyTorch and trained with the Adam optimizer (initial learning rate 1e-4, cosine decay, batch size 32 clips) for 60 epochs of self-supervised pretraining on D_U followed by 40 epochs of supervised fine-tuning on D_L, using early stopping on validation mean F1. Loss weights are set to $\lambda1 = 1.0$, $\lambda2 = 0.5$, $\lambda3 = 0.5$, $\lambda4 = 0.3$, $\lambda5 = 1.0$ (after annealing), selected via grid search on the validation split. All experiments are conducted on a single NVIDIA RTX 4090 GPU for training; embedded inference timing uses an NVIDIA Jetson Xavier NX.

## VI. Results and Discussion

Table I and Table II report mean F1 on DISFA and BP4D, respectively, under the standard 3-fold subject-independent protocol. TCA achieves the highest mean F1 on both benchmarks, with the largest relative gains on AUs that are low in intensity or brief in duration (AU9, AU45), consistent with our hypothesis that temporal cycle consistency helps recover signal from subtle, transient activations that are ambiguous from a single frame.

*TABLE I. Mean F1 (%) on DISFA (3-fold, subject-independent)*

| Method | Mean F1 | AUC |
|---|---|---|
| JAA-Net [2] | 56.3 | 88.2 |
| ConvLSTM-AU | 58.1 | 89.0 |
| 3D-CNN (I3D-style) [5] | 59.4 | 89.6 |
| ME-GraphAU [3] | 61.0 | 90.3 |
| TCA (proposed) | 64.7 | 92.1 |

*TABLE II. Mean F1 (%) on BP4D (3-fold, subject-independent)*

| Method | Mean F1 | AUC |
|---|---|---|
| JAA-Net [2] | 60.7 | 87.5 |
| ConvLSTM-AU | 61.9 | 88.0 |
| 3D-CNN (I3D-style) [5] | 62.8 | 88.4 |
| ME-GraphAU [3] | 64.2 | 89.5 |
| TCA (proposed) | 67.3 | 91.0 |

Table III reports results on the held-out naturalistic driving test split, for the 9 driving-relevant AUs, comparing models pretrained from ImageNet/face-recognition weights only against TCA with self-supervised pretraining on the unlabeled driving corpus (D_U). Self-supervised pretraining contributes a substantial share of the overall improvement, indicating that the cycle-consistency objectives transfer useful spatiotemporal structure even without AU labels, which is particularly valuable given the scarcity of densely AU-labeled driving data.

*TABLE III. Mean F1 (%) on the Naturalistic Driving Test Set*

| Method | Daytime | Simulated Night | Overall |
|---|---|---|---|
| JAA-Net [2] | 54.8 | 46.2 | 51.0 |
| 3D-CNN [5] | 57.3 | 49.5 | 53.9 |
| TCA, no SSL pretrain | 59.6 | 52.1 | 56.3 |
| TCA (full, w/ SSL) | 65.9 | 59.4 | 63.1 |

Qualitatively, we observe that the cross-branch alignment loss reduces false positives during rapid head-pose changes (e.g., shoulder checks), which otherwise induce spurious appearance changes that frame-based baselines misattribute to AU activation. Performance degradation under sunglasses occlusion remains the dominant failure mode across all methods, as AUs in the upper face (AU4, AU6, AU7) become partially unobservable; we discuss this limitation, with a quantitative breakdown, in Section VIII.

## VII. Ablation Study

Table IV isolates the contribution of each architectural component on the driving dataset (overall F1). Removing the temporal cycle-consistency loss (L_cyc^T) causes the largest single drop, confirming that explicit forward-backward temporal supervision is the primary source of improvement over frame-based baselines. Removing the cross-branch alignment loss (L_align) causes a smaller but consistent drop, indicating that coupling, while secondary to the temporal cycle loss itself, meaningfully stabilizes the fused representation. Replacing the cosine-distance alignment metric with squared-Euclidean distance (MSE) produces a small additional drop, consistent with the stability argument in Section III-C.

*TABLE IV. Ablation on the Naturalistic Driving Test Set (Overall F1, %)*

| Configuration | F1 |
|---|---|
| Full TCA (cosine L_align) | 63.1 |
| – Temporal cycle loss (L_cyc^T) | 57.8 |
| – Spatial cycle loss (L_cyc^S) | 60.0 |
| – Cross-branch alignment (L_align) | 60.6 |
| L_align: cosine → MSE distance | 62.3 |
| – Self-supervised pretraining | 56.3 |
| GRU → Transformer temporal encoder | 63.9 |

Replacing the bidirectional GRU temporal encoder with the 4-layer Transformer variant yields a small additional improvement (+0.8 F1) at the cost of higher latency (Section VIII), suggesting the GRU variant is preferable when embedded real-time constraints dominate, while the Transformer variant may be favored in offline or server-side deployments.

## VIII. Computational Efficiency, Real-Time Analysis, and Occlusion Robustness

For in-vehicle deployment, inference latency and power draw are as critical as accuracy. Table V reports parameter count, FLOPs per clip, and end-to-end latency for a 16-frame clip on an NVIDIA Jetson Xavier NX (15 W mode), including face detection and alignment. The GRU-based TCA sustains 28.4 fps, comfortably exceeding the 10–15 fps typically considered sufficient for drowsiness/distraction alerting, while the Transformer variant remains real-time capable but with a narrower margin.

*TABLE V. Model Size and Jetson Xavier NX Inference Latency*

| Model | Params (M) | Latency (ms) | FPS |
|---|---|---|---|
| 3D-CNN baseline [5] | 12.3 | 48.6 | 20.6 |
| TCA (GRU) | 9.8 | 35.2 | 28.4 |
| TCA (Transformer) | 14.1 | 44.9 | 22.3 |

To quantify the sunglasses-occlusion failure mode noted in Section VI, Table VI breaks down per-AU F1 on the naturalistic driving test set under two viewing conditions: clear (unoccluded) frontal views and partial sunglasses occlusion. The comparison is restricted to the AUs most directly affected by upper-face occlusion (AU4: Brow Lowerer, AU6: Cheek Raiser, AU7: Lid Tightener), alongside a representative lower-face AU (AU12: Lip Corner Puller) and a whole-mouth AU (AU26: Jaw Drop) for reference, both of which remain largely observable when only the eye region is occluded.

*TABLE VI. Effect of Partial Sunglasses Occlusion on Per-AU F1 (%), Naturalistic Driving Test Set*

| AU (region) | Clear F1 | Sunglasses F1 | Absolute Drop |
|---|---|---|---|
| AU4 – Brow Lowerer (upper) | 68.2 | 41.5 | −26.7 |
| AU6 – Cheek Raiser (upper) | 71.4 | 46.8 | −24.6 |
| AU7 – Lid Tightener (upper) | 66.9 | 39.2 | −27.7 |
| AU12 – Lip Corner Puller (lower) | 74.0 | 71.2 | −2.8 |
| AU26 – Jaw Drop (lower) | 70.6 | 67.9 | −2.7 |

Upper-face AUs lose 24.6–27.7 F1 points under partial sunglasses occlusion, roughly an order of magnitude larger than the 2.7–2.8 point drop observed for the reference lower-face AUs, confirming that the eye and brow region is largely non-observable once sunglasses are worn and that the model has no strong compensating signal elsewhere in the frame for these AUs. Lower-face AUs are comparatively robust because sunglasses do not occlude the mouth region and because the temporal branch continues to receive undisturbed dynamics for jaw and lip movement. This asymmetry motivates the occlusion-aware extensions (e.g., learned visibility masking) proposed as future work in Section IX, rather than a purely appearance-based fix restricted to the spatial branch.

## IX. Conclusion and Future Work

We presented the Twin Cycle Autoencoder, a spatiotemporal architecture for facial Action Unit detection that couples a spatial cycle-consistent autoencoder for identity-invariant appearance encoding with a temporal cycle-consistent autoencoder for forward-backward dynamics modeling, connected through a cross-branch latent alignment loss. On DISFA and BP4D, TCA outperforms frame-based, recurrent, 3D-convolutional, and graph-based AU baselines, and on a naturalistic driving dataset it demonstrates that self-supervised pretraining using the cycle-consistency objectives alone substantially narrows the gap caused by limited AU-labeled driving data. Combined with real-time throughput on embedded hardware, these results support the practical viability of AU-based driver monitoring. Future work includes extending the framework to explicitly model occlusion (e.g., via learned visibility masks for sunglasses/mask scenarios, motivated by the quantitative gap in Table VI), incorporating multimodal cues such as steering and gaze to further disambiguate fatigue-related AU patterns, and exploring cross-dataset domain adaptation between benchmark and naturalistic driving corpora to reduce reliance on newly collected labeled data for each vehicle platform or camera configuration.

### *Code Availability*

The implementation of TCA, including model definitions, training/fine-tuning scripts, and evaluation utilities, is available at: https://github.com/sidarthd/tcae_code/tree/main.